\documentclass[letterpaper, 10 pt, conference]{ieeeconf}  
\usepackage{tabularx}
\usepackage{multirow}

\IEEEoverridecommandlockouts                              

\usepackage{booktabs}
\usepackage{graphics} 
\usepackage{amsmath} 
\usepackage{amssymb}  
\usepackage[colorlinks, linkcolor=blue, citecolor=blue]{hyperref}
\usepackage{amssymb}
\usepackage{graphicx}
\usepackage{caption}
\usepackage{subfigure}
\usepackage{mathrsfs}
\usepackage{array}
\usepackage{multirow}
\usepackage{makecell}
\usepackage{xcolor}
\usepackage{amsfonts}
\usepackage{gensymb}

\usepackage{biblatex}
\usepackage{flushend}

\title{\LARGE \bf
CORAL: Colored structural representation for \\bi-modal place recognition
}

\author{Yiyuan Pan, Xuecheng Xu, Weijie Li, Yunxiang Cui, Yue Wang, Rong Xiong
\thanks{Yiyuan Pan, Xuecheng Xu, Weijie Li, Yunxiang Cui, Yue Wang, and Rong Xiong are with the State Key Laboratory of Industrial Control Technology and Institute of Cyber-Systems and Control, Zhejiang University, Zhejiang, China. Yue Wang is the corresponding author {\tt\small wangyue@iipc.zju.edu.cn}.}
}

\begin{document}

\maketitle
\thispagestyle{empty}
\pagestyle{empty}

\begin{abstract}
	
Place recognition is indispensable for a drift-free localization system. Due to the variations of the environment, place recognition using single-modality has limitations. In this paper, we propose a bi-modal place recognition method, which can extract a compound global descriptor from the two modalities, vision and LiDAR. Specifically, we first build the elevation image generated from 3D points as a structural representation. Then, we derive the correspondences between 3D points and image pixels that are further used in merging the pixel-wise visual features into the elevation map grids. In this way, we fuse the structural features and visual features in the consistent bird-eye view frame, yielding a semantic representation, namely CORAL. And the whole network is called CORAL-VLAD. Comparisons on the Oxford RobotCar show that CORAL-VLAD has superior performance against other state-of-the-art methods. We also demonstrate that our network can be generalized to other scenes and sensor configurations on cross-city datasets.
\end{abstract}


\section{INTRODUCTION}

Loop closure is essential for the localization system because of the drift reduction, especially in large-scale outdoor scenes. A popular pipeline for loop closure usually employs place recognition as the first step, since it is able to find a place from the large map database that is close to the current place, based on purely sensor data similarity. Therefore, place recognition is widely applied in various autonomous robot systems for navigation.

The camera is the most popular sensor as it observes the texture of the environment at a low cost. Therefore, visual place recognition draws research attention for years. A traditional pipeline is to build a global feature descriptor by aggregating handcrafted sparse local features for each image, e.g. SIFT \cite{lowe2004distinctive} and SURF \cite{bay2008speeded}. Then the efficient searching method is used to look for the nearest global descriptor on the database as the most similar match. With the development of the deep neural network, convolution neural networks (CNNs) demonstrate promising retrieval performance \cite{arandjelovic2016netvlad, noh2017large,tang2020adversarial} by extracting more reliable image descriptors. However, due to the susceptibility of the visual image under strong variations of seasons, weather, illumination, and viewpoints, building robust and discriminative image features remains a challenge.

\begin{figure}[tp]
	\centering
	\includegraphics[width=0.98\linewidth]{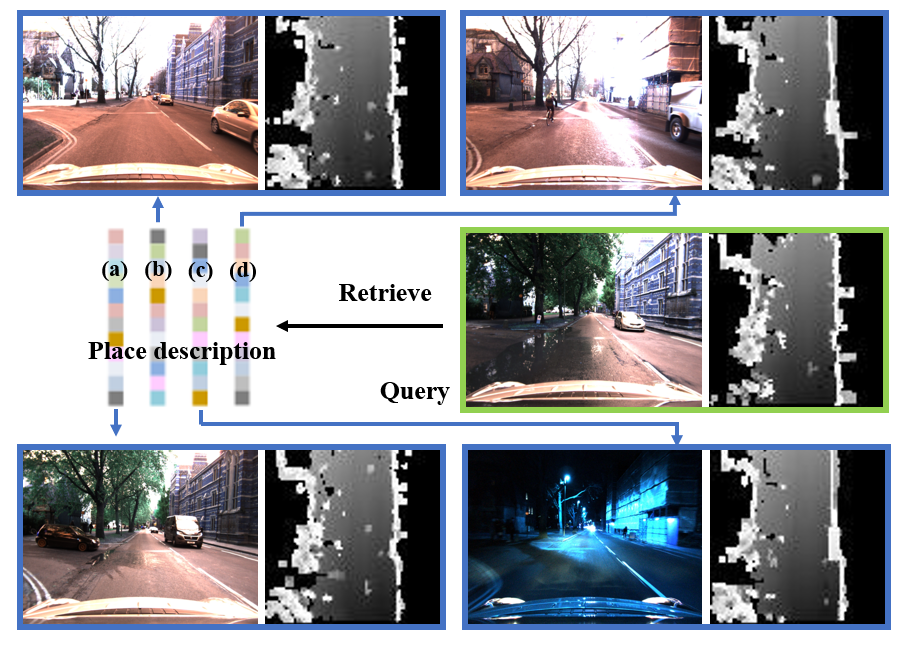}
	
	\caption{ The proposed bi-modal coupling of representation fusing visual image and elevation image retrieves the most similar sample in the reference maps with different environmental conditions, like (a) overcast, (b) sun, (c) night, and (d) dawn.}
	\label{environment}
	\vspace{-10pt}
\end{figure}

To relieve the problem, LiDAR has been an alternative that provides accurate and relatively stable 3D structural information. Following the visual place recognition pipeline, LiDAR-based place recognition methods employ deep neural networks to extract structural features from LiDAR scans \cite{liu2019lpd, angelina2018pointnetvlad}, demonstrating superior performance than the visual place recognition, especially in changing outdoor environment. However, LiDAR still has its weakness when the environmental structure has fewer features.

Due to the inherent shortages of both the camera and LiDAR, it is difficult to extract appropriate features using the single-modality sensor to describe numerous complex scenes. Thus, multi-sensor data fusion is becoming a feasible solution for place recognition. A common sensor for fusion image and geometry information is the RGB-D camera but it's not credible in the outdoor scenes. Using the camera and LiDAR is a more robust way to deal with multi-sensor fusion. Whereas, inconsistent viewpoints, diverse observation ranges and data structures of the different sensor data become major issues to generate a compound global descriptor effectively. In existing fusion-based place recognition methods, the visual and structural information are processed independently to generate their own descriptors and then concatenated directly as a compound descriptor without consistency in geometry.

In this paper, we set to combine the two modalities, vision and LiDAR, for place recognition, as shown in Fig. \ref{environment}. The main novelty is the construction of the bi-modal fused representation, namely colored structural representation (CORAL). As shown in previous works on LiDAR place recognition, various 3D representations, including point cloud \cite{angelina2018pointnetvlad}, histogram \cite{yin2018locnet}, and polar image \cite{gkim-2019-ral,xu2020disco}, are designed to represent LiDAR scans, which highly impact the efficiency and effectiveness. We propose to build a local dense elevation map to describe the environmental structure. The map also derives the correspondences between 3D points and image pixels. The pixel-wise visual features are then inserted into the elevation map grids to semantically `colorize' the structural features. With such tightly bi-modal coupling, CORAL encodes both visual and structural features in the same consistent bird-eye view (BEV) frame. The contributions can be summarized as:
\begin{itemize}
\item A local dense elevation map representation is utilized for place recognition, which is irrelevant to LiDAR hardware configurations to encode the structural information.
\item A semantic representation with corresponding geometry named CORAL is proposed for place recognition which is robust towards various environmental changes.	
\item A validation on experiments is conducted to evaluate the performance of the proposed method, which shows superior performance in testing and generalization datasets. The code is also released \footnote[1]{https://github.com/Panyiyuan96/CORAL\_Pytorch.git}.
\end{itemize}

%
%
%

\section{Related Work}

\textbf{Vision-based place recognition} Vision-based place recognition is typically regarded as the problem of image retrieval which is solved by searching the most similar match on the reference database. Traditionally, some salient image parts are encoded as handcrafted local features, such as SURF \cite{bay2008speeded}, SIFT \cite{lowe2004distinctive}, or ORB \cite{mur2015orb, mur2017orb}, and then these local features can be aggregated to a global descriptor leveraging feature aggregated methods, such as Bag-of-visual-words \cite{galvez2012bags}, VLAD \cite{jegou2010aggregating} and Fisher Vectors \cite{jegou2011aggregating}. Obtaining the global descriptor of a query, efficient searching methods like KD-tree search can be used to look for the closest global descriptor as the most similar match on the reference database. In recent years, handcrafted local features have been increasingly replaced by learnable features using deep neural networks that have significant improvement in extracting descriptive descriptors. Several mature networks for extracting local features, like VGG-Net \cite{simonyan2014very} and ResNet \cite{he2016deep}, achieve an amazing performance on place recognition. As for aggregating local features, inspired by the traditional method - VLAD, NetVLAD \cite{arandjelovic2016netvlad} is formulated as a learnable function that is better in local features clustering. Generalized-Mean Pooling \cite{radenovic2018fine} is also an efficient and differentiable aggregated method, allows the network to capture a compact global descriptor in an end-to-end fashion.
	
\textbf{Structural-based place recognition} Considering structural features are more robust in changing environments, structural-based methods become an alternative for place recognition. Handcrafted structural descriptors, like PFH \cite{rusu2008aligning} and SHOT \cite{salti2014shot}, usually have poor generalizability, that they can only be used in speciﬁc tasks. To relieve this problem, convolution neural networks are used to extract structural descriptors. Owing to orderless of point clouds, several works convert the raw point clouds into a 3D volume representation, such as VoxelNet \cite{zhou2018voxelnet} and volumetric CNNs \cite{qi2016volumetric}. However, the conversion process introduces high quantization loss and requires lots of computation time. PointNet \cite{qi2017pointnet} is a pioneering work that is able to capture structural features from raw point clouds directly. Combining PointNet and NetVLAD, PointNetVLAD \cite{angelina2018pointnetvlad} is the first approach to achieve large-scale long-term place recognition with the input of raw point data. However, PointNet operates each point independently thus ignoring the local structure relationship of points. In response to this problem, LPD-Net \cite{liu2019lpd} proposes the adaptive local feature extraction module and the graph-based neighborhood aggregation module. PCAN \cite{zhang2019pcan} introduces an attention map to predict the significance of local regions. Both methods boost the performance of place recognition.

\textbf{Fusion-based place recognition} Multi-sensor fusion solution by incorporating visual and structural features is a more appropriate way to achieve place recognition under strong environment variations. In recent years, a few fusion-based works have emerged for place recognition. The naive strategy \cite{xie2020large, oertel2020augmenting} of generating a compound descriptor from different modality inputs is directly combining two global descriptors extracted from two independent network streams, ignoring the inner relationship of visual context and geometry structure in the same region. Referring to some 3D object detection methods, works of \cite{wang2018fusing, liang2018deep} fuse the visual and structural information by operating the intermediate structural feature map and visual feature map in the same frame, which inspires us to construct a more discriminative and robust compound descriptor.  

\begin{figure*}[tp]
	\centering
	\vspace{3mm}
	\includegraphics[width=0.95\textwidth]{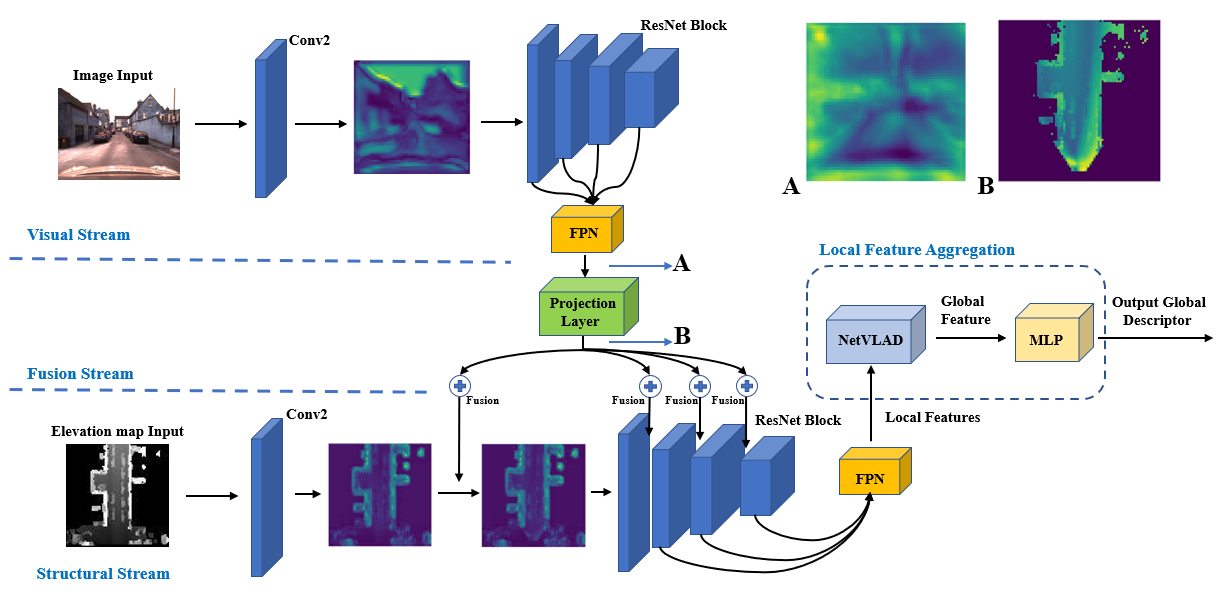}
	
	\caption{The network architecture of CORAL-VLAD. The network inputs an elevation image generated by LiDAR scans and a visual image, and then outputs a global descriptor to predict the most similar scene on the reference database. $A$ shows multi-scale visual feature map from the output of FPN layer, $B$ shows the BEV visual feature map from the ouput of projection layer.}
	\label{architecture}
	\vspace{-10pt}
\end{figure*}

\section{Methodology}

The architecture of the proposed fusion network is shown in Fig. \ref{architecture}. The network inputs consist of two streams: raw front-view images from the camera and filtered elevation images generated from the LiDAR scans. The two-stream network tightly couples the visual and structural features which enforce the final representation to encode the semantics into the geometric structure. Specifically, the dense elevation image representation encodes the structural information, and it also provides the point-to-pixel correspondences, which is leveraged to insert visual features into the consistent BEV frame. Therefore, aggregating the bi-modal features is geometrically sensible, yielding the final global descriptor for place recognition.

\subsection{Elevation map generation}

We introduce the elevation map representation defined on a grid map. Each 2D grid indexes an elevation to describe the environment structure. Originated in the 2D occupancy grid map, the elevation map replaces the occupancy information with the elevation, which is capable of representing the 2.5D ground surface.

First, the 6 degree-of-freedom (DOF) pose of the sensor denoted as $T$ is calculated by LiDAR inertial odometry algorithm, with respect to the global frame $G$. Given a 3D point $p_i$ of a new measurement in the sensor frame, we transform it into the global frame by $T p_i$ and calculate the elevation:
\begin{equation}\label{eq1}
e_p = P T p_i
\end{equation}
the projection matrix $P = [0,0,1]$ retrieves 3rd entry of the transformed point as the elevation $e_p$, while the first two entries are rounded and then converted to the corresponding index of the grid map. For processing multiple observations of the same grid, the variance $\sigma_{p}$ of elevation is introduced to describe the uncertainty of elevation measurement according to range sensor models \cite{fankhauser2015kinect}. In this way, each grid data $({e_g}, \sigma_{g}^{2})$ is updated according to the corresponding measurement data $({e_p}, \sigma_{p}^{2})$. Note that only the measurement within the Mahalanobis distance threshold of the grid is fused by variance weighted strategy to obtain an updated grid data as follows:
\begin{equation}\label{eq3}
{e_g} = \frac{\sigma_{p}^{2}{e_g}+\sigma_{g}^{2}e_p}{\sigma_{p}^{2}+\sigma_{g}^{2}} \qquad
\sigma_{g}^{2} = \frac{\sigma_{p}^{2}\sigma_{g}^{2}}{\sigma_{p}^{2}+\sigma_{g}^{2}}
\end{equation}
Furthermore, to clear the dynamic objects due to the use of the accumulation mechanism, ray tracing is utilized to check whether the grid is crossed by a ray, and if the grid is occupied by a dynamic object, the elevation and variance of this grid are initialized. The detailed generation process can be found in \cite{pan2020gem}.

\begin{figure}[tp]
	\centering
	\includegraphics[width=0.98\linewidth]{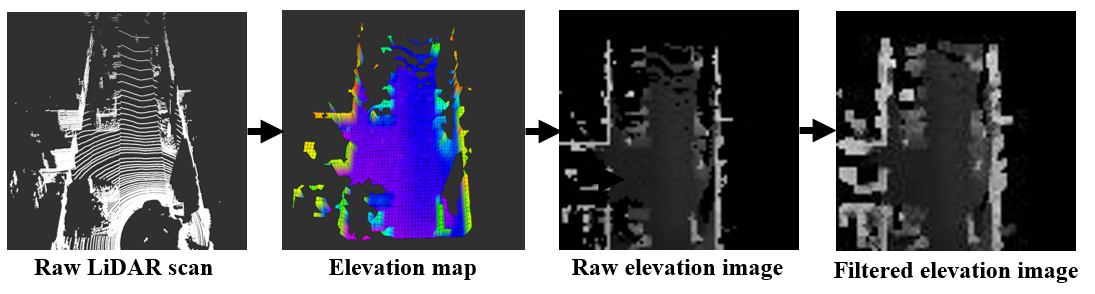}
	
	\caption{Elevation image generation. The elevation map accumulated from raw LiDAR scans is projected on the image plane as a raw elevation image. The filtered elevation image is generated using a mean filter.}
	\label{ele_generation}
	\vspace{-10pt}
\end{figure}

\textbf{Elevation image} To utilize 2D convolution neural networks directly, the elevation map is converted into a one-channel grayscale image with the same size as the grid map, namely elevation image. As shown in Fig. \ref{ele_generation}, we scale the elevation value from $0$ to $255$ as the grayscale value of the elevation image. Additionally, a mean filter is applied to fill up some invalid pixels which have not been observed on the elevation image using a $3\times3$ filter template. Compared with the single scan based BEV representation, the elevation image is much denser to describe the structure, also leading to more accurate point-to-pixel correspondences.

\subsection{Network architecture}

The architecture of the proposed compound network is shown in Fig. \ref{architecture}. The whole network can be divided into three main components: the feature extraction module, the fusion layer, and the local feature aggregation module.

\textbf{Feature extraction} Our multi-sensor network has two streams for feature extraction. The visual stream employs the lightweight ResNet18 \cite{he2016deep} as the backbone to efficiently extract the visual features. Due to the utilize of several convolutional layers with stride 2, the output feature map suffers a dramatic reduction in size which is not suitable for fusion. So we use feature pyramid network (FPN) to combine four feature maps from the residual block group to recover the final feature map with the size same as the first residual block and exploit multi-scale feature information. 

The structural stream comprises a group of convolutional layers to capture structural features, and four groups of residual blocks to extract fusion features. Except for the first group, all groups start with the convolution layer with stride 2 and all other convolutions are with stride 1. The number of $3\times3$ kernel convolutions in each group is 2, 4, 4, 6, 6, and each group outputs the feature vector with the corresponding dimensions of 64, 64, 128, 192, and 256 respectively. The outputs of the last three residual blocks also exploit multi-scale information as the visual stream does.

\textbf{Fusion Layer} The fusion layer comprises two components: the projection layer and the fusion module. The goal of the projection layer is to convert the front-view visual feature map into the BEV visual feature map through sparse matrix multiplication. The grid index of the elevation image and the corresponding elevation value can be converted into a 3D position in the global frame. So each grid can be denoted as a 3D point and transformed into the LiDAR frame according to the estimated pose. With the intrinsic parameters of the camera and extrinsic parameters between the LiDAR and the camera, each 3D point can be projected onto the 2D camera image plane which helps retrieve corresponding visual image features. Following this idea, we implement a parameterless projection layer to find the correspondences. Considering that the coordinates of 2D projected points are often non-integers, we combine the visual features from adjacent discrete pixels by bilinear interpolating.  After that visual feature map is generated in the BEV frame consistent with the elevation image, thus achieving appropriate features corresponding to later fusion.

Then, we arrive at the generation of CORAL representation in the fusion module. Denote the $i$th layer of structural feature map as $S_i$, and the corresponding BEV visual feature map as $V_i$. To maintain the same shape of $S_i$ and $V_i$, pooling operation $Pool(\centerdot)$ is used to adjust the output size of the raw projection layer $V$ and $1\times1$ kernel convolution $Conv(\centerdot)$ is applied to keep the same channel size. We try two different methods for aggregating features. The first one uses element-wise concatenation to obtain CORAL $F_i$ defined by
\begin{equation}\label{eq3}
F_i = \left[Conv(Pool(V)), S_i\right]
\end{equation}
In the second method, feature maps are combined by element-wise summation given by
\begin{equation}\label{eq3}
F_i = Conv(Pool(V)) + S_i
\end{equation}

We also propose two fusion strategies that combine the structural feature map and the visual feature map from the first residual layer, or all four blocks, which are evaluated in Section IV.

\textbf{Local feature aggregation} The local feature aggregation module learns to further extract a global descriptor. Considering the capacity of the NetVLAD on aggregating features, we feed CORAL to a NetVLAD layer and generate a global descriptor. Furthermore, we utilize a multi-layer perception (MLP) to process the raw output of NetVLAD for learning a dimension reduction mapping to decrease the size of the global descriptor which accelerates the nearest neighbor search.

\subsection{Training the fusion descriptor}

We train our compound network in an end-to-end fashion to yield a bi-modal fused global descriptor. A margin-based loss is adopted to train pairs of samples labeled with positive or negative based on corresponding GPS position.

\textbf{Loss function} The training data is constructed as sets of tuples $ \mathcal{T} = (P_a, \left\{P_{pos}\right\}, \left\{ P_{neg}\right\})$. $P_a$ denotes an anchor with a pair of visual image and elevation image, $\left\{P_{pos}\right\}, \left\{ P_{neg}\right\}$ represent the set of anchor's positive matches and negative matches determined by the their relative distances. We apply the squared Euclidean distance $\delta$ to evaluate the similarity of two global descriptors. Margin-based loss is used to minimize the distance $\delta_{a,pos}$ between the global descriptors of matching samples $(P_a, P_{pos})$ while pushing apart the dissimilar matches $(P_a, P_{neg})$. To achieve faster convergence and better discrimination, we only use the closest/hardest negative $P^{-}_{neg}$ in $\left\{P_{neg}\right\}$ and the most dissimilar positive $P^{+}_{pos}$ in $\left\{P_{pos}\right\}$ during back propagation. Comparing with various retrival loss functions, we utilize the lazy quadruplet \cite{chen2017beyond} as the training loss:
\begin{equation}\label{eq3}
\begin{split}
\mathcal{L}(\mathcal{T}, P_{neg*}) = max([\alpha + \delta_{a,pos} - \delta_{a,neg})]_{+})\\
+max([\beta + \delta_{a,pos} - \delta_{a,neg*}]_{+})
\end{split}
\end{equation}
where $[\centerdot]_+$ is the hinge loss, $\alpha, \beta$ are constant margin parameters, and $P_{neg*}$ is randomly sampled from training data which is disimilar to all observations of $\mathcal{T}$.

\textbf{Data sampling strategy} The original hard-negative training strategy usually offers faster convergence but it may lead to a collapsed model. To alleviate this problem, we divide the training process into two stages. In the first stage, negative matches $\left\{ P_{neg}\right\}$ are sampled randomly from all negative samples. This is followed by a second training stage, negative matches $\left\{ P_{neg}\right\}$ are generated by looking for the hardest pairs from all global descriptors of negatives calculated by the latest model. This hard-negative mining strategy ensures negative samples become progressively harder, which avoids prolonged convergence and boosts the performance of the converged model.

\section{Experiments}

In this section, we discuss the datasets and settings for training and evaluation. Quantitative results are provided to demonstrate the performance of our method on diverse scene conditions and cross-city generalization experiments. Furthermore, the loss parameters are set as $\alpha=0.5$, $\beta=0.2$, the number of positive matches $\left\{P_{pos}\right\}$ is 2 and negative matches $\left\{P_{neg}\right\}$ is 18.

\begin{center}
	\begin{table}[tp]
		\vspace{1.5mm}
		\caption{Comparison results with the average recall@1 and recall@1\% of different networks on the Oxford dataset.}
		\centering
		\linespread{1.1}\selectfont
		\begin{tabularx}{7.1cm}{c|c|c}
			\toprule[1.5pt]
			& Ave recall@1 & Ave recall@1\%\\ [1pt]
			\hline
			PN-VLAD & 67.94 & 81.01\\[1pt]
			LPD-Net & 86.28 & 94.42\\[1pt]
			Img-VLAD& 64.47 &85.24\\[1pt]
			Aug-Net& 79.47  &91.24\\[1pt]
			\hline
			Vis-VLAD(ours)& 57.62 & 83.05 \\[1pt]
			Ele-VLAD(ours)& 82.49 & 93.61 \\[1pt]
			Sum-First(ours)& 82.44 & 92.71 \\[1pt]
			Con-First(ours)& 84.82 & 93.62    \\[1pt]
			Sum-Four(ours)& 86.23 & 94.43\\[1pt]
			Con-Four(ours)& \textbf{88.93} & \textbf{96.13}   \\[1pt]
			\toprule[1.5pt]
		\end{tabularx}
		\vspace{-12pt}
		\label{performance}
	\end{table}
\end{center}

\subsection{Datasets and settings}

We choose the Oxford Robotcar dataset \cite{maddern20171} for training and testing. The car equipped with LiDAR sensors and cameras repeatedly drives through the same regions in one year and collects data in large-scale and long-term conditions across weather, season and illumination. To create training tuples, we define the similar observations with a relative distance less than $10m$ and heading difference less than 30 degrees as positive pairs, and those at least $50m$ apart as negative pairs. Referring to the data splitting rules of \cite{angelina2018pointnetvlad}, we obtain 21636 training tuples from 44 sequences of the original dataset and 3011 testing samples from 23 sequences. To keep large overlapping regions between the camera image and the elevation image, the elevation map is set with the size of $80\times80$ and the resolution of $0.5m$. The input of the visual image is downscaled into $112\times112$ for extracting visual features efficiently. Furthermore, as scans are accumulated by 2D LiDAR scans at fixed intervals, there are some incomplete samples on the dataset. We discard them when generating the elevation image.

\textbf{Generalization settings} We choose KITTI dataset \cite{geiger2012we} to evaluate the generalization performance across city and sensor configurations. Only Sequence 00 which visits the same places repeatedly is used. The first $170s$ of Sequence 00 are used to construct the reference map and the rest are used as localization queries. The second dataset for generalization evaluation is collected by ourselves in different seasons, namely YQ. Data in overcast scene and snowy scene are used for mapping and query respectively. For a fair comparison, the same criteria on Oxford Robotcar dataset are applied.

\begin{figure}[tp]
	\centering
	\includegraphics[width=0.98\linewidth]{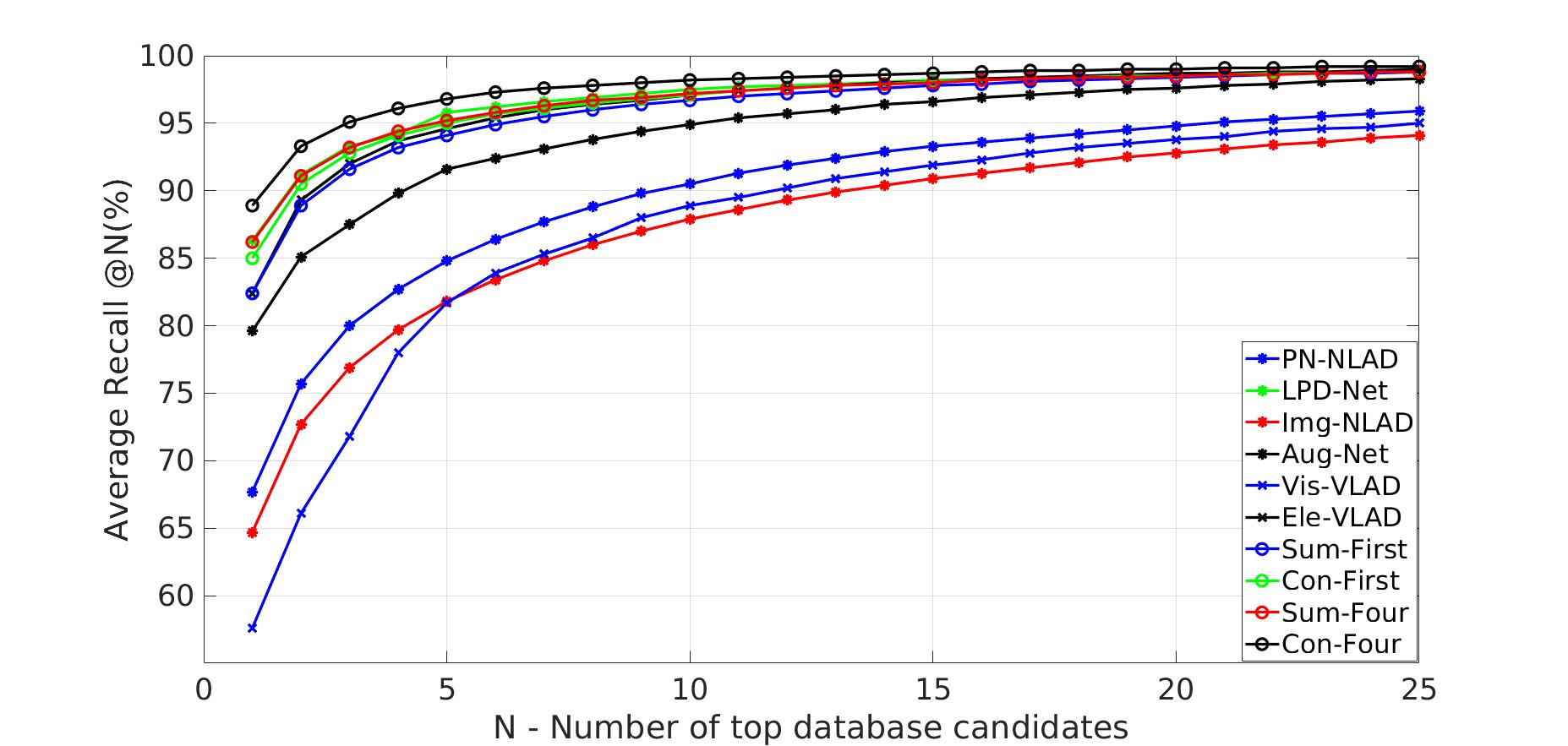}
	
	\caption{Average recall@N(\%) of different networks on the Oxford dataset.}
	\label{recall1}
	\vspace{-10pt}
\end{figure}

\begin{figure}[tp]
	\centering
	\vspace{10pt}
	\includegraphics[width=0.98\linewidth]{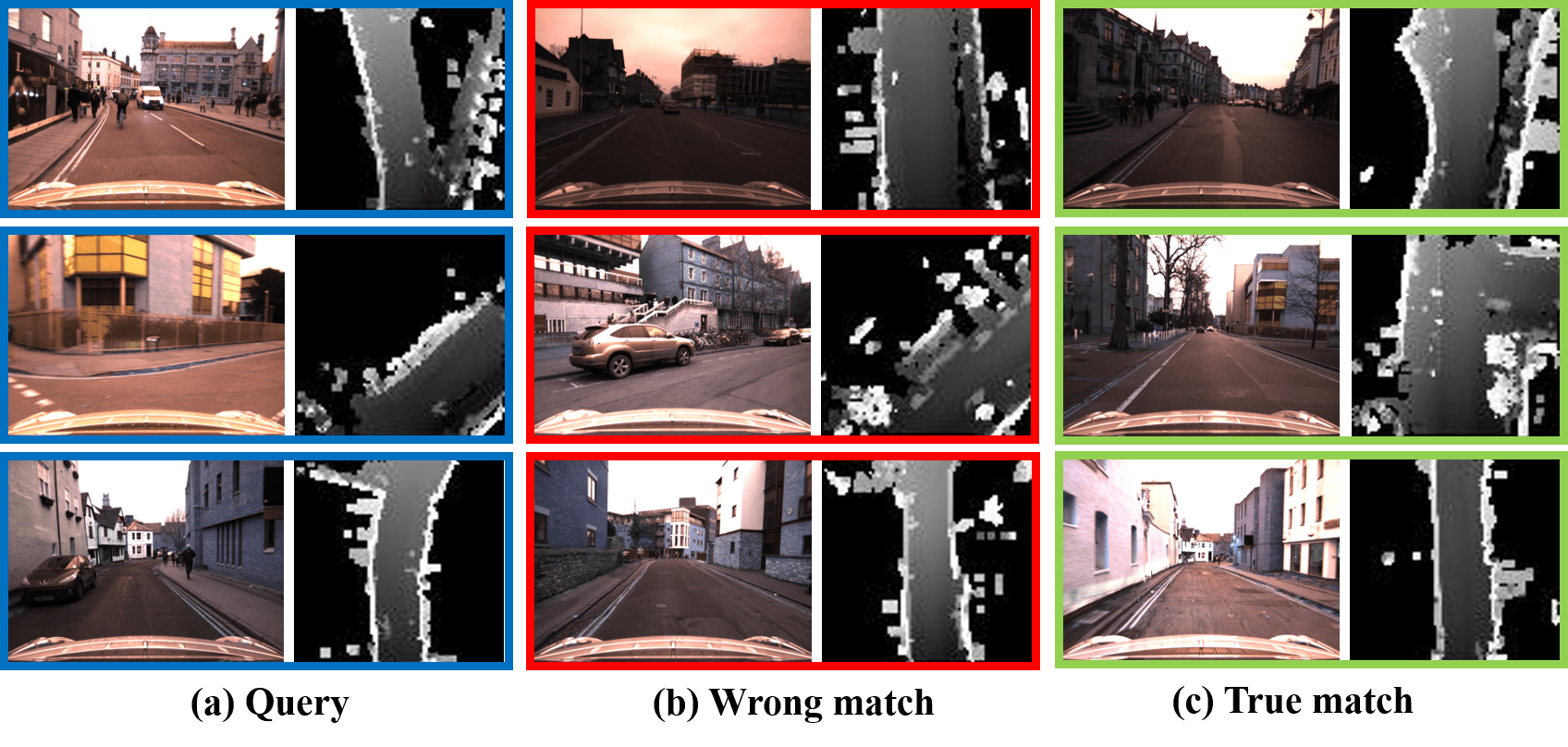}
	
	\caption{Networks limitations. These are incorrect matches retrieved by our network, where (a) is the query, (b) is the wrong match, and (c) shows the true match. }
	\label{limitation}
	\vspace{-6pt}
\end{figure}

\begin{figure*}[tp]
	\centering
	\vspace{1.5mm}
	\includegraphics[width=1\textwidth]{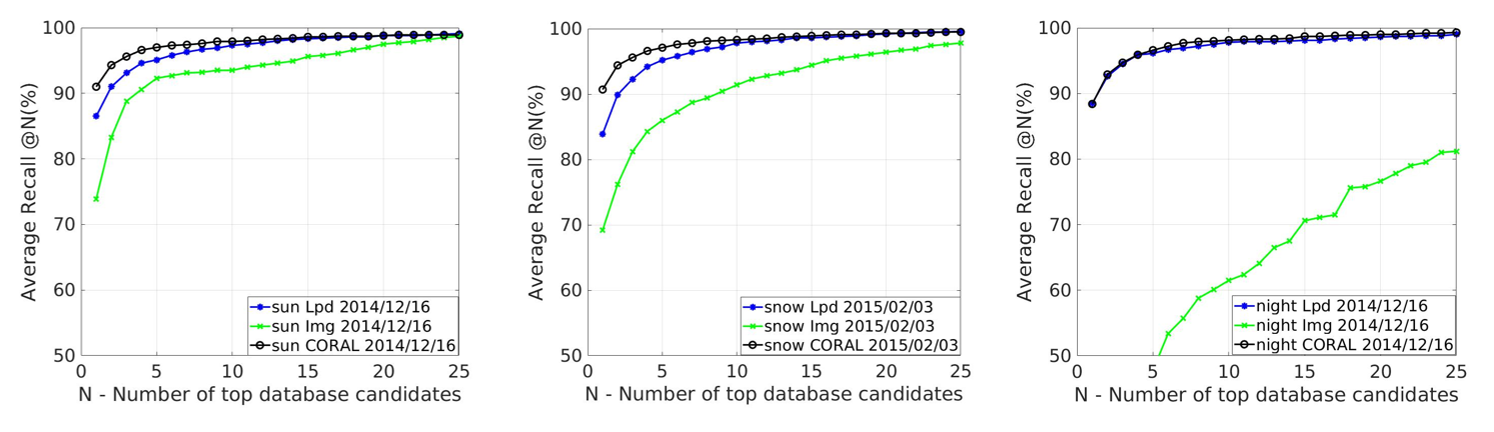}
	
	\caption{Average recall@N(\%) with LPD-Net(Lpd), Img-VLAD(Img) and CORAL-VLAD(CORAL) under different scene conditions on the Oxford dataset.}
	\label{condition_recall}
	\vspace{-10pt}
\end{figure*}

\begin{center}
	\begin{table}[tp]
		\caption{Average timing for computing a single instance using NVIDIA 2080 Ti.}
		\centering
		\linespread{1.1}\selectfont
		\begin{tabularx}{8cm}{cc}
			\toprule[1.5pt]
			Network & Avg. computational cost per descriptor\\ [1pt]
			\hline
			PN-VLAD & 8.44ms \\[1pt]
			LPD-Net & 15.28ms \\[1pt]
			Img-NetVLAD & 62.28ms \\[1pt]
			Aug-Net & 18.80ms \\[1pt]
			\hline
			CORAL-VLAD(ours)& 11.19ms \\[1pt]
			\toprule[1.5pt]
		\end{tabularx}
		
		\label{time}
		\vspace{-12pt}
	\end{table}
\end{center}

\subsection{Place recognition results on the Oxford dataset}

We present qualitative results to demonstrate the feasibility of our CORAL-VLAD on the Oxford dataset under changing scene conditions. The common assessment indicators of place recognition - average recall@1 and the average recall@1\% are used to evaluate the network performance. For fairness, the final global descriptor dimensions of all networks are set to 256.

\textbf{Ablation study on fusion strategy} We test our compound network with four feature fusion strategies operating on the intermediate visual feature map and the structural feature map, including (a) element-wise summarization in the first residual block (Sum-First), (b) element-wise concatenation in the first residual block (Con-First), (c) element-wise summarization in four residual blocks (Sum-Four), (d) element-wise concatenation in four residual blocks (Con-Four). The results are shown in Fig. \ref{recall1} and Tab. \ref{performance}, the multi-scale fusion strategies outperform the single-scale due to the adequate information. And the better performance of concatenation than summation contributes to the intact information. In the following experiments, we use our network with the Con-Four fusion strategy as the final version, denoted as CORAL-VLAD.

\textbf{Ablation study on sensor modal} Furthermore, to investigate the contribution of the fusion step, we separate our fusion architecture into an independent visual stream and structural stream with a shared NetVLAD module to generate the global descriptor, called Vision NetVLAD(Vis-VLAD) and Elevation image NetVLAD(Ele-VLAD). The results are also shown in Fig. \ref{recall1} and Tab. \ref{performance}. Except for the fusion strategy of Sum-First, results have been significantly improved using composite features. Furthermore, the results for single modal place recognition are shown, validating the correct design and implementation of our method. 

\textbf{Comparison with the-state-of-art methods}  We compare our method with the state-of-the-art single modal place recognition methods, PointNetVLAD (PN-VLAD) \cite{angelina2018pointnetvlad}, LPD-Net \cite{liu2019lpd} and Img-VLAD \cite{arandjelovic2016netvlad}, as well as bi-modal place recognition method, compound network (Aug-Net)\cite{oertel2020augmenting}. 

Using only structural data input, the performance of our Ele-VLAD is better than PN-VLAD, yet slightly worse than LPD-Net. Inputs of LPD-Net and PN-VLAD are 4096 filtered points with detailed 3D structural information of the environment while the elevation image only has $40\times40$ grids with one-channel elevation. It also proves effectiveness of the elevation image for representing 3D geometric data. Although providing a rich appearance context, Vis-VLAD and Img-VLAD cannot outperform Ele-VLAD, which proves that the elevation image representation is more robust under environment variations. Results are shown in Fig. \ref{environment}. Meanwhile, there are many over-exposed images on the Oxford dataset, causing loss of visual features. 

Using composite descriptors, the performance of CORAL-VLAD exceeds all other methods, including Aug-Net. Note that there is a difference in the Aug-Net from \cite{oertel2020augmenting}, since we employ the PN-VLAD splitting rules clarified in \cite{angelina2018pointnetvlad} to ensure consistency among all methods. All comparison network training sessions last no longer than $24h$. We suppose that the main reason for the slightly worse results of Aug-Net is the incompleteness of the 3D volume based representation.

\textbf{Comparison under different conditions}
We compare the performance of LPD-Net, Img-VLAD, CORAL-VLAD, which use inputs in different representation modalities under changing scene conditions. Fig. \ref{condition_recall} shows the results when queries are taken from three different scene conditions against the testing database on the Oxford database (except query run). It can be seen that the huge variants of environmental conditions have little impact on retrieval performance using structural cues. As almost all corresponding visual features are lost at night scene, the visual-based approach obtains lots of incorrect matches. Accordingly, CORAL-VLAD has minimal improvement over LPD-Net in such condition, showing the limitation of additional visual modal.

\textbf{Cases study} Fig. \ref{environment} shows some of the successfully matched results in changing environments. We can observe that our network has learned robust features and alleviated the negative effect brought by dynamic obstacles and variations in illumination. Fig. \ref{limitation} shows three wrong cases. We can see that the network is confused in the same scenes with opposite viewpoints (top row) and a few overlapping areas (middle row and bottom row).

\textbf{Comparison on efficiency} We further evaluate the running time of the network. Due to the use of lightweight network backbones and compact structural representation, our network implementation takes about $11ms$ which is faster than all methods except PN-VLAD shown in Tab. \ref{time}. For generating the elevation map, we have a GPU-based implementation at almost 30Hz as shown in \cite{pan2019gpu}, achieving real-time performance for robotics applications.

\begin{center}
	\begin{table}[tp]
		\vspace{1.5mm}
		\caption{Comparison generalization results of the average recall@1(\%) on the KITTI and YQ dataset}
		\centering
		\linespread{1.1}\selectfont
		\begin{tabularx}{7.9cm}{c|c|c|c}
			\toprule[1.5pt]
			Network& KITTI\_laser & KITTI\_stereo & YQ \\
			\hline
			PN-VLAD& 72.43 &65.43 & 40.36 \\[1pt]
			LPD-Net & 74.58 & 65.82 & 62.35 \\[1pt]
			Aug-Net&75.60 &70.56 & 66.91 \\[1pt]
			\hline
			CORAL-VLAD(ours)& \textbf{76.43} & \textbf{70.77} & \textbf{73.82}  \\[1pt]
			\toprule[1.5pt]
		\end{tabularx}
		
		\label{Generalizability}
	\vspace{-10pt}	
	\end{table}
\end{center}

\subsection{Generalization evaluation}

To analyze the generalization of our network, we evaluate our network on the YQ and KITTI datasets using the model trained on the Oxford Robotcar dataset. These cross-city datasets consist of unobserved conditions with different sensor configurations, including sensor types and extrinsic parameters. Specifically, KITTI\_laser uses the point cloud collected by LiDAR while KITTI\_stereo uses the point cloud generated from stereo images.  

\begin{figure}[tp]
	\vspace{2.6mm}
	\centering
	\includegraphics[width=0.98\linewidth]{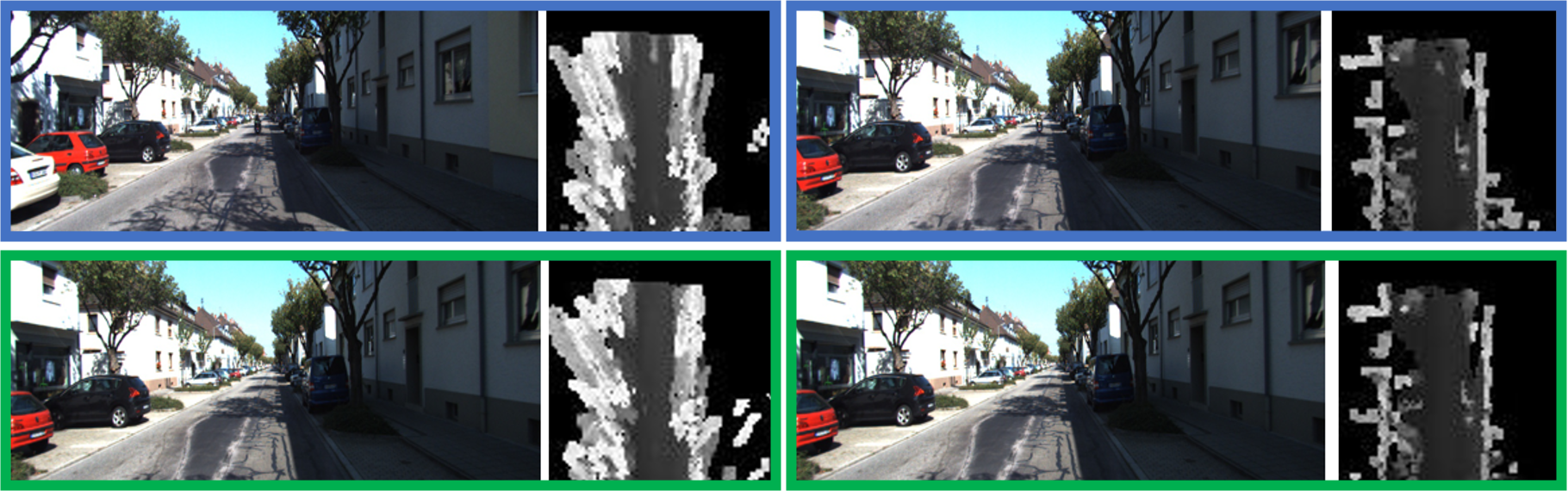}
	
	\caption{Matching examples on the KITTI dataset. The left column shows the correct match on the KITTI\_stereo and the right column shows the correct match on the KITTI\_laser.}
	\label{Kitti}
	\vspace{-3pt}
	
\end{figure}

In the case of KITTI\_laser dataset, the elevation image is generated in real-time using Velodyne 64 HDL LiDAR shown in Fig. \ref{Kitti} and our network outperforms other methods in Tab. \ref{Generalizability}. Although there is noise involved during the calculation of the disparity map on the KITTI\_stereo dataset, generated elevation maps become blurred, which loses some structural information and introduces inaccurate matches between pixels and 3D points. CORAL-VLAD still achieves the best results, demonstrating the validity of our network for pure vision-based applications. 

\begin{figure}[tp]
	\centering
	\includegraphics[width=0.98\linewidth]{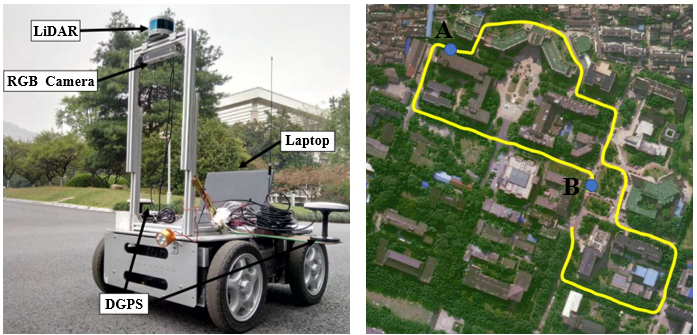}
	
	\caption{The collection of the YQ dataset. The left image shows the data collection platform. The yellow line of the right image shows the trajectory of the data collection on our campus. }
	\label{YQ_p}
	\vspace{-10pt}
\end{figure}

\begin{figure}[tp]
    \vspace{2.6mm}
	\centering
	\includegraphics[width=0.98\linewidth]{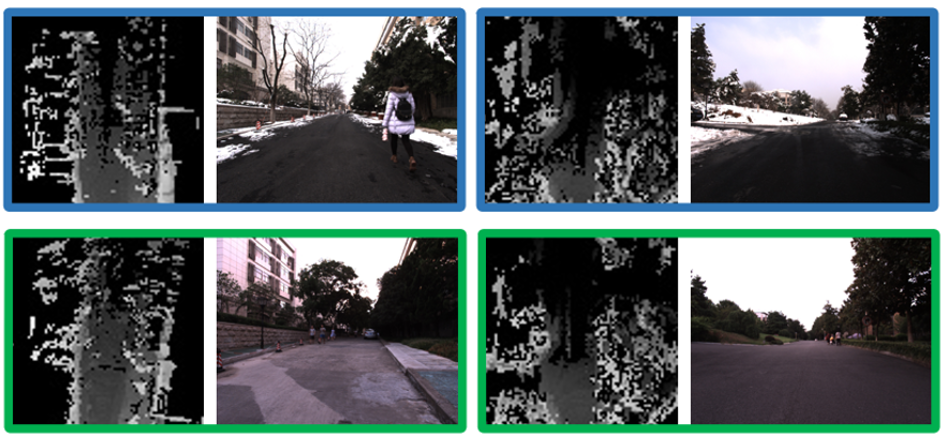}
	
	\caption{Matching examples on the YQ dataset. The left column shows the correct matching example at point A in Fig. \ref{YQ_p}. The right column shows the matching result at point B in Fig. \ref{YQ_p}.}
	\label{YQ}
	\vspace{-10pt}
\end{figure}

Additionally, we conduct experiments under different weather conditions on our campus. Unlike evaluating with similar conditions at a short period of time on the KITTI dataset, the unmanned ground vehicle equipped with a LiDAR and an RGB camera collects sensor data on sunny days in spring and snowy days in winter denoted as YQ dataset. The data collection platform is shown in Fig. \ref{YQ_p}. In addition, DGPS provides the ground-truth position to check the correctness of the matching results. The summer-overcast scene of the YQ dataset is utilized as the reference database, and the winter-snow scene as the query dataset. Matching examples are shown in Fig. \ref{YQ}, formulating a generalization evaluation on both conditions and sensor configurations. CORAL-VLAD still achieves the best performance in this scenario as shown in Tab. \ref{Generalizability}, because of the elevation image based structural representation, transforming the texture into a geometric sensible representation.

\section{Conclusion}

In this paper, we introduce the elevation map as the structural information and propose the bi-modal environment representation CORAL to fuse the structural and visual features in the same consistent BEV frame, which can handle various environmental variances like viewpoint changes, illuminations, and structure losses. We show that the method performs best on the Oxford Robotcar dataset, as well as generalization test on conditions and sensor configurations using the cross-city datasets of the KITTI dataset and YQ dataset.

\section{Acknowledgment}

This work was supported in part by the National Nature Science Foundation of China under Grant 61903332, and in part by the Natural Science Foundation of Zhejiang Province under grant number LGG21F030012.

\printbibliography

\end{document}